# Toward a Benchmark for Controllable Simulation of Imperfect Students with Large Language Models

Alexander Apartsin[a], Omri Sason[b], Yehudit Aperstein[b]
[a]Holon Institute of Technology, Holon, Israel.
[b]Afeka Tel Aviv Academic College of Engineering, Tel Aviv, Israel

**Abstract**

Teacher education requires deliberate practice with learners who exhibit identifiable strengths, weaknesses, and partial mastery. Large language models could support such practice by simulating students with known skill components, enabling teachers to rehearse explanations, diagnoses, and instructional responses. For this purpose, however, the central requirement is neither to maximize benchmark accuracy nor to suppress isolated facts, but to control model behavior so that it reflects a specified skill profile.

This paper investigates whether prompted language models can be steered to retain some skills while suppressing others. We introduce a benchmark-oriented framework in which an explicit skill vector represents a simulated student, prompt-based control specifies retained and missing competencies, and behavior is evaluated using profile-alignment metrics, retained-versus-forgotten comparisons, and cross-skill calibration analyses.

The results show that selective partial mastery can be induced and measured in a structured mathematics setting, although the degree of controllability remains model-dependent. These findings position controllable learner simulation as a distinct research problem at the intersection of teacher education, educational simulation, and language-model control.

## 1. Introduction

The main motivation of this work is to use large language models to simulate imperfect students for teacher deliberate practice. In such settings, teachers benefit from repeated interactions with learners whose strengths, weaknesses, and partial mastery are known in advance, allowing them to practice explanation, diagnosis, and instructional response under controlled conditions. This creates a practical requirement for learner simulators: they must behave like students with specified skill profiles rather than collapsing into uniformly capable assistants. We therefore study *controllable skill profiles*, meaning the ability to stably and measurably preserve some skills while suppressing others. The present paper addresses the narrower benchmark problem rather than the broader problem of building a fully pedagogical virtual student.

### 1.1 Motivation: The Teacher Training Challenge

Teacher preparation is a setting in which controlled student simulation could be especially valuable. The transition from academic preparation to classroom practice is a formative phase in which novice teachers face heterogeneous learner needs and must adapt instruction in real time (Ingersoll & Strong, 2011). Deliberate-practice research further suggests that expertise develops through repeated engagement with targeted challenges rather than generic exposure (Ericsson et al., 1993). For teacher development, this means practice with learners who display partial mastery, uneven reasoning, and domain-specific difficulty in ways that can be revisited and compared.

However, authentic classroom experience is difficult to standardize. A trainee may encounter a particular difficulty with fractions but never revisit the same instructional pattern under comparable conditions. Computational simulation can help by enabling repeated, controlled exposure to varied learner profiles (Heffernan & Heffernan, 2014). For that purpose, a useful simulator must exhibit the omissions and weaknesses that the instructional scenario is meant to surface, and it should do so in ways that remain interpretable to the teacher.

The key requirement is therefore not generic imperfection, but *controllable partial mastery*. A simulator that remains uniformly expert, regardless of the requested learner profile, is poorly matched to this goal. At the same time, answer-level control alone is insufficient to ensure realistic teacher training. In this paper, we focus on the narrower, more measurable prerequisite: can an LLM be directed to behave like a student with a specified skill profile by retaining some curricular skills while suppressing others, with minimal collateral damage?

### 1.2 The Fundamental Challenge: Why Controllability Differs from Performance

Current evaluation paradigms for LLMs center on task performance: higher benchmark accuracy is treated as a measure of better model quality. That framing has driven substantial progress, but it creates a misalignment when the goal shifts from maximization to specification. In educational simulation, we do not simply want a model that performs *well*; we want a model that performs *as specified*, for example, by preserving some competencies while predictably failing on others.

This distinction reveals a central limitation of performance-first evaluation: strong baseline accuracy does not guarantee controllability. A model that solves mathematics well may still fail to simulate a student who has forgotten a specific skill family, especially when the prompt asks it to behave against its stronger pretrained competence. We therefore evaluate not only what the models know, but how selectively their displayed competencies can be modulated.

We introduce **controllability** as a distinct evaluation dimension orthogonal to traditional performance metrics. For educational simulation, controllability means the ability to realize specified patterns of knowledge retention and forgetting rather than merely demonstrating high raw accuracy.

### 1.3 The Skill vs. Fact Distinction: Why Skill Forgetting is Harder

Understanding controllability for educational simulation requires distinguishing between two fundamentally different targets of control: **fact forgetting** and **skill-profile control**. This distinction is critical for both evaluation methodology and practical implementation.

**Fact forgetting**, as explored in machine unlearning research (Bourtoule et al., 2021), refers to the removal of discrete factual assertions from a model's behavior. Facts have clear boundaries and unambiguous truth conditions: "Paris is the capital of France" is either asserted or it is not. Evaluating fact forgetting is straightforward; the model either correctly declines to state the targeted fact or fails to do so. Recent advances in machine unlearning and LLM unlearning (Xu et al., 2023; Ji et al., 2024; Guo et al., 2024) have produced increasingly effective methods for targeted knowledge removal.

**Skill-profile control** operates at a fundamentally different level of abstraction. Skills represent generalizable competencies, sets of related abilities enabling correct performance across families of related tasks. The skill "solving linear equations" encompasses numerous specific problem types, each requiring appropriate application of underlying principles to different contexts. Unlike facts, skills lack crisp definitional boundaries. The question is therefore not simply whether a model can avoid one answer, but whether it can sustain a coherent pattern of partial mastery across related items.

This boundary ambiguity introduces measurement challenges that fact forgetting does not face. Skill mastery exists along a continuum, manifesting differently across varied problem types. A student with weak fraction knowledge might still solve some fraction problems correctly, depending on problem structure, difficulty, and specific misconceptions. Unlike fact forgetting, where complete success means never stating the fact, skill-profile control requires navigating a space of partial and systematic behavior that still reflects the intended knowledge configuration.

Furthermore, skills exhibit hierarchical relationships and dependencies. Weakening one skill family may have differential effects on related competencies. The structured nature of controllable partial mastery, ideally involving interpretable omissions rather than random mistakes, requires evaluation approaches that capture not just accuracy reduction but behavioral pattern alignment.

### 1.4 Additional Challenges: Why Prompt-Based Control is Non-Trivial

Achieving controllable skill-profile behavior through prompt-based instructions faces additional challenges that distinguish this task from conventional prompt engineering.

**Deep knowledge entrenchment.** Contemporary LLMs acquire knowledge through extensive pretraining on vast corpora of human-generated text. This process creates deeply entrenched internal representations that may resist modification through contextual cues alone. When explicitly instructed to behave like a learner with missing skills, models may continue applying their stronger internal competence due to the strength of pretraining effects.

**Interference effects.** Skills are not isolated but interconnected through shared underlying concepts and reasoning strategies. Suppressing one skill might induce unintended degradation in related competencies, a phenomenon analogous to catastrophic interference in neural network training. Conversely, instructions intended to weaken one component might inadvertently reinforce related knowledge through emphasis or repetition.

**Evaluation methodology.** Traditional accuracy-based metrics fail to capture the nuanced nature of controllable partial mastery. A model achieving 50% accuracy on a skill domain might be exhibiting the intended suppression, might find the material inherently challenging, or might be applying a flawed strategy that coincidentally produces correct answers. Disentangling these possibilities requires careful baseline comparison and skill-level analysis of response patterns.

**Model resistance.** Models trained with strong instruction-following objectives may resist behavioral modifications that conflict with their internalized representations. Understanding which architectures and training paradigms facilitate or impede controllability is essential for practical deployment.

### 1.5 Our Approach: Controllable Generative Students

This work introduces a benchmark-oriented framework for **controllable generative students** in large language models. The framework combines explicit skill-vector profiles, prompt-based control, and evaluation targeted at selective retention and suppression rather than raw correctness alone. The resulting manuscript should be read as an investigation of controllable skill profiles in a mathematics benchmark, with teacher training as a motivating application rather than a completed deployment claim.

### 1.6 Research Questions and Contributions

**Central Research Question:** Can we configure a prompted LLM to simulate a student with a specified skill profile by selectively retaining some skills while suppressing others?

This investigation addresses three sub-questions: (1) How effective are different prompting strategies at enforcing skill-level behavioral control? (2) How does controllability vary across model architectures? (3) Can specified skill profiles be realized with limited interference across skills?

**Contributions**

We make four primary contributions:

1. **A benchmark formulation for partial mastery control** based on binary skill vectors that specify retained and forgotten skills explicitly;
2. **A prompt-based student-profile mechanism** for testing whether LLM behavior can be steered toward selective retention and suppression;
3. **An evaluation framework** combining RMSE, retained-versus-forgotten summaries, and calibration diagnostics to separate controllability from baseline capability;
4. **A comparative empirical study** across prompting strategies and model families in a curriculum-aligned mathematics setting.

The remainder of this paper is organized as follows. Section 2 reviews related work on teacher training, educational simulation, behavioral control, and knowledge tracing. Section 3 formalizes the problem and introduces the evaluation metrics. Section 4 describes the dataset. Section 5 presents the methodology. Section 6 summarizes the concrete experimental setup used in the reported study. Section 7 reports the experimental results. Section 8 discusses implications and limitations. Section 9 concludes with future directions.

## 2. Related Work

### 2.1 Teacher Professional Development and Deliberate Practice

Teacher preparation research emphasizes the difficulty of moving from coursework into real classroom practice. Induction and mentoring studies show that novice teachers benefit from structured support, guided reflection, and repeated opportunities to interpret learner behavior under authentic constraints (Ingersoll & Strong, 2011). From this perspective, simulated learners are useful only if they expose trainees to meaningful variability rather than to uniformly correct answers.

Ericsson's deliberate-practice framework (Ericsson et al., 1993) further motivates repeated practice with targeted challenges rather than generic repetition for teacher development, thereby involving encounters with students exhibiting partial mastery, uneven progress, and domain-specific difficulty under conditions that can be revisited and compared. Intelligent tutoring environments such as ASSISTments illustrate the broader value of controlled educational interaction platforms, even though they were not designed as LLM-based student simulators (Heffernan & Heffernan, 2014).

More recently, Aperstein, Cohen, and Apartsin (2025) review generative-AI platforms for deliberate teaching practice and argue that simulated classroom environments can support repeated, goal-oriented teacher rehearsal. That perspective is closely aligned with the motivation of the present study. It highlights that a useful training platform depends on controllable learner behavior rather than on generic text generation alone.

This literature motivates our application setting, but it also defines an important boundary. Educational relevance does not follow automatically from answer-level control. A controllable benchmark can support future teacher-training simulations, yet additional work is still needed to validate dialogue quality, pedagogical realism, and fidelity to misconceptions.

### 2.2 LLM-Based Educational Simulation

Recent work shows that LLMs can participate in educational simulations and tutoring workflows. However, the emphasis is usually on dialogue quality, personalization, or broad profile conditioning rather than on tightly measured skill-level control. Broader reviews of LLMs in education likewise emphasize tutoring, feedback, assessment, and personalization as dominant application modes (Gan et al., 2023). Within that landscape, Zhang et al. (2025) demonstrate classroom simulation with LLM-powered agents, and Lu and Wang (2024) introduce generative students represented through binary mastery vectors.

Other systems focus on tutoring behavior rather than simulated student controllability. AI2T (Weitekamp et al., 2024), SocraticLM (Liu et al., 2024), and tutoring-oriented systems such as Scarlatos et al. (2025) and Guevarra et al. (2025) study how LLMs can support instruction, scaffolding, or interaction design. These works are closely related, but they do not primarily ask whether a model can be forced to preserve certain curricular skills while suppressing others in a tightly defined benchmark.

Our work, therefore, fits within LLM-based educational simulation while taking a narrower target: not full tutoring quality or open-ended student dialogue, but benchmarked control of partial mastery patterns in answer behavior.

### 2.3 Persona and Behavioral Control in LLMs

Persona and behavior-control research is relevant because it studies whether prompted LLMs can maintain non-default behavioral states over extended interactions. PersonaLLM (Jiang et al., 2023), RoleLLM (Wang et al., 2023), PersonaGym (Samuel et al., 2024), and recent role-playing surveys (Tseng et al., 2024) all show that behavioral conditioning is possible but imperfect. Fidelity depends on the model, the target behavior, and the evaluation protocol.

Related work also studies how prompts and backstories can stabilize role enactment. Moon et al. (2024) construct virtual personas through curated backstory anthologies, while Kadavath et al. (2022) show that LLMs expose metacognitive signals relevant to calibration and self-assessment. More broadly, prompting research treats prompts as a general control surface rather than merely as few-shot examples (Liu et al., 2023), and inference-time steering methods such as PASTA show that stronger behavioral control can also be imposed by directing model attention to user-specified information without parameter updates (Zhang et al., 2024). Prompt-design surveys further suggest that multi-step and iterative prompting can improve control over complex tasks by decomposing constraints across stages (Cohen & Aperstein, 2024). Together, these papers motivate treating persona control, confidence-aware behavior, and prompt structure as useful neighboring problems, even though our target is curricular competence rather than narrative identity.

Our problem differs from persona consistency in one important way: the target is not a style or identity trait but a curricular competence profile. The relevant success condition is not narrative coherence or voice stability, but selective degradation on targeted skills with limited spillover to untargeted ones.

### 2.4 Knowledge Removal and Behavioral Control

Knowledge-removal and machine-unlearning research asks a closely related question: can model behavior be altered selectively while preserving non-target behavior? Classical machine-unlearning work (Bourtoule et al., 2021; Xu et al., 2023) and more recent LLM-specific methods (Ji et al., 2024; Guo et al., 2024) focus on the forget-retain trade-off and on limiting collateral damage to retained knowledge.

Our setting differs from that literature in two ways. First, we use prompting rather than training-time unlearning interventions. Second, the target is not the removal of a discrete set of facts, but the suppression of a curricular skill family expressed across many related items. The forget-retain perspective is still useful, however, because it sharpens the central benchmark question: can targeted suppression be achieved without disproportionate loss on retained skills?

### 2.5 Knowledge Tracing and Learner Modeling

Knowledge tracing and learner modeling provide the closest educational analogue to our notion of a skill profile. Deep Knowledge Tracing (Piech et al., 2015), Dynamic Key-Value Memory Networks (Zhang et al., 2017), and self-attentive tracing (Pandey & Karypis, 2019) all model learner state over time. Gervet et al. (2020) further show that model choice in knowledge tracing depends strongly on data regime and task structure, while EdNet (Choi et al., 2020) illustrates the scale and hierarchy that modern learner-modeling datasets can support.

The difference is that knowledge tracing aims to infer or predict a learner's latent state from observed interactions. Our paper instead specifies an external target state and asks whether a generative model can be steered to behave as if that state were present. In that sense, the present work is complementary to knowledge tracing: tracing estimates mastery, whereas our benchmark probes whether predefined mastery patterns can be rendered controllably.

### 2.6 Research Gaps and Positioning

Despite substantial progress in educational simulation, persona conditioning, behavioral control, and knowledge tracing, prior work still leaves several central questions unresolved.

**Gap 1: Limited quantitative evaluation of partial mastery control.** Prior educational-simulation work has demonstrated profile conditioning and qualitative realism. However, the cited literature does not yet provide a compact benchmark centered on measured retain-versus-forget behavior at the skill level.

**Gap 2: Weak connection between steering literature and curricular skill control.** Persona and steering studies show that LLM behavior can be nudged, but they do not directly establish whether curricular skills can be suppressed selectively while preserving untargeted competencies.

**Gap 3: Missing bridge between learner-state representation and controllable generation.** Knowledge-tracing research provides representations of mastery, yet it is not designed to test whether a model can be instructed to enact a specified mastery vector in a controlled benchmark.

We address these gaps by introducing a benchmark framework for controllable imperfect-student simulation that uses explicit skill vectors and an evaluation focused on selective retention, targeted suppression, and collateral effects across skills. Our claim is therefore narrower than “realistic student simulation” in general: we study whether prompt-based control can produce measurable partial-mastery behavior in a curriculum-aligned mathematics benchmark.

## 3. Problem Formulation

### 3.1 Formal Definitions

Let $\mathcal{S} = \{s_1, \dots, s_K\}$ denote the set of skill domains for a given grade, and let $\boldsymbol{k} = (k_1, \dots, k_K) \in \{0,1\}^K$ denote the target mastery vector:

- $k_i = 1$: Student has **retained** skill $s_i$

- $k_i = 0$: Student has **forgotten** skill $s_i$

For each skill $s_i$, let $a_i^0 = A_{\text{base}}(s_i)$ denote the baseline accuracy under the perfect-student configuration and let $a_i^* = A_{\text{ctrl}}(s_i \mid \boldsymbol{k})$ denote the observed accuracy under profile $\boldsymbol{k}$. Thus, $\boldsymbol{k}_{\text{perfect}} = (1, \ldots, 1)$ represents full mastery, whereas any vector containing zeros represents an imperfect student with targeted skill gaps.

For a given profile, let $R = \{i: k_i = 1\}$ be the retained skill set and $F = \{i: k_i = 0\}$ be the forgotten skill set. We summarize aggregate retained and forgotten behavior with

$$\bar{a}_R(\boldsymbol{k}) = \frac{1}{|R|}\sum_{i \in R} a_i^*, \qquad \bar{a}_F(\boldsymbol{k}) = \frac{1}{|F|}\sum_{i \in F} a_i^*$$

These quantities support the retained-versus-forgotten comparisons used throughout the Results section. The main text emphasizes RMSE, retained-versus-forgotten separation, and calibration diagnostics because those are the most interpretable quantities across all analyzed grades and models.

## 3.2 Evaluation Metrics

### 3.2.1 Relative Loss

Relative loss measures the proportional drop in performance relative to the perfect-student baseline:

$$\mathcal{L}_i = \max\left(0, \frac{a_i^0 - a_i^*}{\max(a_i^0, \varepsilon)}\right)$$

Here $\varepsilon > 0$ is a small constant preventing division by zero. For forgotten skills, larger values are desirable because they indicate stronger suppression. For retained skills, values near zero are desirable because they indicate preserved performance.

### 3.2.2 Root Mean Squared Error

$$\text{RMSE}(\boldsymbol{k}) = \sqrt{\frac{1}{K}\sum_{i=1}^{K}(k_i - a_i^*)^2}$$

RMSE measures the discrepancy between the intended binary profile and the observed continuous performance profile. Lower values indicate closer alignment between specification and behavior.

### 3.2.3 Controllability Score

$$C(\boldsymbol{k}) = \frac{1}{K}\sum_{i=1}^{K}[k_i a_i^* + (1 - k_i)(1 - a_i^*)]$$

This score rewards high accuracy on retained skills and low accuracy on forgotten skills. It lies in [0,1], with $C = 1$ indicating perfect agreement with the target profile.

### 3.2.4 Cross-Skill Influence

To assess interference across skills, we define the cross-skill influence matrix

$$\Delta_{ij} = A_{\text{ctrl}}(s_j \mid k_i = 0) - A_{\text{base}}(s_j)$$

where $\Delta_{ii}$ captures the direct effect of suppressing $s_i$ and $\Delta_{ij}$ for $i \neq j$ captures collateral effects on other skills. In the main results, this dependence is summarized through correlation structure and calibration figures rather than through a full matrix table, because the preserved evidence is easier to interpret at that level.

#### 3.2.5 Prediction Score

$$\mathrm{PS}(\boldsymbol{k}) = \bar{a}_R^{\text{actual}}(\boldsymbol{k}) - \bar{a}_R^{\text{expected}}(\boldsymbol{k})$$

Here $R = \{i: k_i = 1\}$ is the set of retained skills, $\bar{a}_R^{\text{actual}}$ is the observed mean retained-skill accuracy for a profile, and $\bar{a}_R^{\text{expected}}$ is the retained-skill mean predicted from the calibration model in Section 5.3. A score near zero indicates agreement with the theoretical expectation; a positive score indicates better-than-expected retention; and a negative score indicates underperformance.

## 4. Dataset

### 4.1 MathCAMPS

The empirical framework uses MathCAMPS: curriculum-aligned mathematics problems tagged to U.S. Common Core standards. Each standard maps to a domain treated as a distinct skill, enabling direct transformation from standards to skill taxonomy. This approach extends prior educational datasets such as EdNet (Choi et al., 2020), which established large-scale hierarchical benchmarks for knowledge tracing in mathematics education.

Eight grade-level datasets (grades 1-8) were created, each containing 100 multiple-choice questions balanced across skill domains. Questions use a four-option format (A-D) with numerically close or conceptually plausible distractors. The benchmark spans grades 1-8, while the detailed controllability analyses shown in the main results focus on Grades 4 and 5.

### 4.2 Dataset Statistics

The MathCAMPS dataset is organized into eight grade-level datasets corresponding to grades 1 through 8. Each grade-specific dataset contains 100 multiple-choice questions balanced across the relevant skill domains for that grade.

#### 4.2.1 Grade-Level Skill Families

Table 1 summarizes the grade-level skill families used to organize the benchmark. Throughout the paper, the term *skill* refers to the curricular unit used for evaluation, operationalized at the domain level in alignment with the Common Core domain structure.

***Table 1:*** *Grade-level skill families used to structure MathCAMPS.*

| Grade | Representative skill families |
|---|---|
| 1 | Operations and Algebraic Thinking; Number and Operations in Base Ten; Measurement and Data; Geometry |
| 2 | Operations and Algebraic Thinking; Number and Operations in Base Ten; Measurement and Data; Geometry |
| 3 | Operations and Algebraic Thinking; Number and Operations in Base Ten; Number and Operations with Fractions; Measurement and Data; Geometry |
| 4 | Operations and Algebraic Thinking; Number and Operations in Base Ten; Number and Operations with Fractions; Measurement and Data; Geometry |
| 5 | Operations and Algebraic Thinking; Number and Operations in Base Ten; Number and Operations with Fractions; Measurement and Data; Geometry |

| Grade | Representative skill families |
|---|---|
| 6 | Ratios and Proportional Relationships; The Number System; Expressions and Equations; Geometry; Statistics and Probability |
| 7 | Ratios and Proportional Relationships; The Number System; Expressions and Equations; Geometry; Statistics and Probability |
| 8 | The Number System; Expressions and Equations; Functions; Geometry; Statistics and Probability |

#### 4.2.2 Example Skill-Annotated Items

Table 2 provides examples of how items were treated analytically. Each row shows a question, its associated skill label, and an example of the multiple-choice conversion procedure used when transforming open-format mathematics items into the four-option evaluation format.

***Table 2:*** *Example item annotations and open-to-multiple-choice conversion examples.*

| Grade | Skill family | Original open-format prompt | Illustrative multiple-choice version |
|---|---|---|---|
| 4 | Number and Operations in Base Ten | Write the value of the digit 7 in 3,742. | A) 7 B) 70 C) 700 D) 7000 |
| 5 | Number and Operations with Fractions | Compute $3/4 + 1/8$. | A) $7/8$ B) $5/8$ C) $1/2$ D) $3/8$ |
| 6 | Ratios and Proportional Relationships | A recipe uses 2 cups of flour for 3 batches. How many cups are needed for 9 batches? | A) 4 B) 6 C) 8 D) 12 |

These examples clarify the annotation logic and the item-format conversion procedure: distractors are selected to be either numerically close to the correct answer or conceptually plausible errors for the targeted skill family.

#### 4.2.3 Quality Control

During preprocessing and evaluation, certain items were identified as systematically ambiguous or poorly constructed because repeated cross-model errors suggested dataset artifacts rather than intended skill gaps. These items were replaced with alternatives from the same educational standard while preserving skill balance.

### 4.3 Per-Skill Mastery

$$\mathrm{Acc}(s_i) = \frac{c_i}{n_i}$$

Here $c_i$ is the number of correct responses on items tagged with skill $s_i$, and $n_i$ is the number of evaluated items for that skill. This quantity is the basic observable used throughout the analysis.

## 5. Methodology

### 5.1 Controllable Generative Student

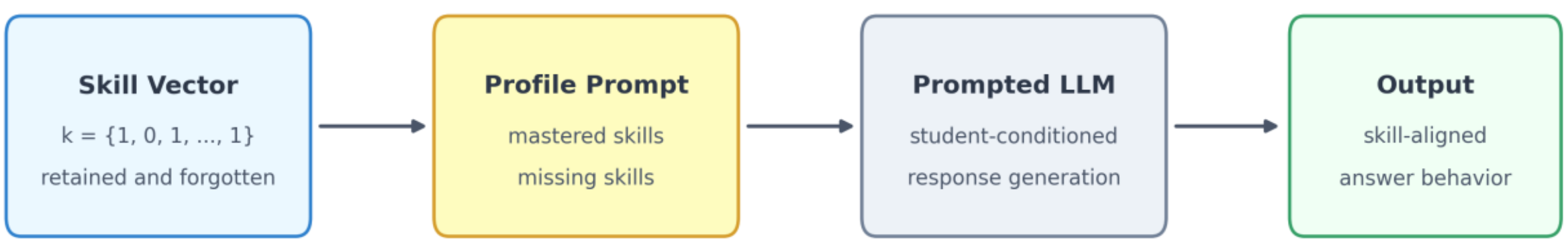

***Figure 1:*** *Controllable Generative Student Architecture. A simulated student is defined by a binary skill vector, where each element corresponds to an evaluated skill. Values of 1 represent retained skills, and values of 0 represent intentionally suppressed skills used to simulate knowledge gaps. The vector is translated into retained and missing skill sets and embedded into a structured student profile prompt that conditions the LLM to produce aligned responses.*

A binary skill vector such as $\boldsymbol{k} = (1,0,\dots,1)$ defines the retained and suppressed skills assigned to a simulated student. Values of 1 indicate retained skills, whereas 0 indicates intentionally suppressed skills. The vector is translated into retained and missing skill sets and embedded into a structured student profile prompt that conditions the LLM to produce aligned responses.

### 5.2 From Skill Vectors to Observable Performance

Evaluation is performed at the student-model pair level and aggregated over all test items. For each skill $s_i$, we compute $\mathrm{Acc}(s_i) = c_i/n_i$, yielding the observed performance vector $\boldsymbol{a}^* = (a_1^*,\dots,a_K^*)$. The profile-level metrics are then obtained by comparing $\boldsymbol{a}^*$ with the target vector $\boldsymbol{k}$. Throughout the Results section, reported ranges summarize variation across the analyzed grades and evaluated student profiles.

Operationally, each simulated student is paired with a skill-tagged question set and produces one answer per item, conditioned on the target mastery vector. These answer-level outputs are then aggregated by skill, profile, and model to form the retained-versus-forgotten comparisons used throughout the paper. This organization allows the same benchmark items to be reused across multiple student profiles while preserving a direct link between the requested skill state and the measured outcome.

### 5.3 Skill Independence and Multivariate Calibration

To study whether forgetting can remain localized, we analyze the dependence structure among skills using per-skill accuracy vectors observed over simulated students. Each student contributes a vector $\boldsymbol{a} = (a_1,\dots,a_K)$, and the Pearson correlation matrix $\Sigma$ summarizes pairwise skill dependencies.

The calibration data are collected at the student level rather than the item level. For a fixed grade and model, one simulated student instance is evaluated over the full skill-tagged question set, and its responses are aggregated into a per-skill accuracy vector $\boldsymbol{a} = (a_1,\dots,a_K)$, where each component is the correctness ratio on items assigned to that skill. Repeating this process over many sampled student instances yields a population of skill-performance vectors from which the empirical mean $\boldsymbol{\mu}$ and covariance $\Sigma$ are estimated. In the underlying calibration design, these repeated student instances are associated with distinct random seeds, so the Gaussian model is intended to capture variation across simulated students rather than variation across individual questions.

**Correlation analysis:** The observed correlations are weak across both grades. Grade 4 values remain close to zero, whereas Grade 5 shows somewhat stronger negative correlations, indicating only mild interference. This pattern is consistent with largely local controllability.

**Multivariate calibration:** As an analytical approximation, we use the first- and second-order moments of the observed skill-performance vectors to estimate expected retained performance under targeted forgetting. Under a Gaussian approximation $\boldsymbol{a} \sim \mathcal{N}(\boldsymbol{\mu}, \Sigma)$, if forgetting is enforced on a subset $F \subset \{1, \dots, K\}$ and the retained set is $R = \{1, \dots, K\} \setminus F$, then the conditional expectation of retained performance is

$$\mathbb{E}[\boldsymbol{a}_R \mid \boldsymbol{a}_F = \boldsymbol{0}] = \boldsymbol{\mu}_R - \Sigma_{RF}\Sigma_{FF}^{-1}\boldsymbol{\mu}_F$$

where $\Sigma_{RF}$ is the cross-covariance block between retained and forgotten skills, and $\Sigma_{FF}$ is the covariance block over forgotten skills. If cross-skill coupling is weak, so that $\Sigma_{RF} \approx 0$, then

$$\mathbb{E}[\boldsymbol{a}_R \mid \boldsymbol{a}_F = \boldsymbol{0}] \approx \boldsymbol{\mu}_R$$

This approximation is used only as a calibration device for later expected-versus-observed comparisons; it is not claimed as a realistic generative model of student cognition.

## 6. Experimental Setup

### 6.1 Models and Prompting Conditions

***Table 3:*** *Models evaluated in this study.*

| Model family | Provider | Selection rationale | Role in study |
|---|---|---|---|
| **Claude** | Anthropic | Strong instruction-following and safety-aligned interaction | Represents a high-compliance frontier model family |
| **DeepSeek** | DeepSeek | Reasoning-oriented open model family with strong baseline math ability | Represents a model family expected to be strong on math but not necessarily on controllability |
| **GPT-4o** | OpenAI | Widely used general-purpose frontier model family | Provides a broadly deployed comparison point |

The reported experiments compare three frontier model families: Claude, DeepSeek, and GPT-4o. All inferences are run with deterministic decoding (temperature = 0), and prompts require the model to return exactly one answer option so that responses can be scored uniformly at scale. The manuscript therefore supports family-level comparison of preserved experimental results rather than a claim about frozen-dated API snapshots.

1. **Instruction-only:** Explicit behavioral constraints instructing correct responses for retained skills and consistent mistakes for suppressed skills.
2. **Example-only:** Demonstration examples illustrating both correct and incorrect behavior. Critically, examples are retrieved from an external file, fully separated from the evaluation datasets, to prevent data leakage.
3. **Hybrid:** Integration of explicit instructions with concrete demonstrations for maximum controllability.

Each prompt follows a consistent architecture containing: (1) the student's retained and missing skills, (2) behavioral instructions and/or examples, and (3) the multiple-choice question. This prompt schema is instantiated programmatically for each student-profile/item pair, enabling systematic comparison across prompt strategies, profiles, grades, and models.

### 6.2 Experimental Protocol and Scope

The experimental workflow has four stages. First, each model is evaluated in a perfect-student condition to establish that the benchmark is solvable without induced forgetting. Second, binary mastery vectors are injected into the prompt to define imperfect students with specific retained and missing skills. Third, the same profiles are tested under instruction-only, example-only, and hybrid prompting to measure the extent to which prompt design affects controllability. Fourth, outputs are aggregated into profile-level and skill-level metrics, including RMSE to the target profile, retained-versus-forgotten performance, forgotten-skill heatmaps, and cross-skill correlation summaries.

The benchmark construction spans grades 1-8, but the detailed controllability analysis reported here focuses on Grades 4 and 5. These grades provide a compact yet nontrivial testbed for exhaustive profile enumeration: Grade 4 supports a four-skill analysis, and Grade 5 supports a three-skill comparative analysis in the preserved results. This keeps the profile space small enough to inspect systematically while still testing whether forgetting can remain selective rather than globally degrading performance. The paper, therefore, reports an in-depth main-text analysis on these grades rather than claiming uniform evidence across the full benchmark.

Each analyzed grade contributes 100 multiple-choice questions, so the main two-grade study operates over 200 item instances in total. All items use a four-option answer format, and each item is assigned to a single skill family for evaluation. This single-skill tagging allows the same question pool to be reused across different simulated student profiles while preserving a clear mapping between intended forgetting and observed per-skill accuracy.

Question construction follows the benchmark procedure described earlier in the paper: open-format mathematics prompts are converted into multiple-choice items, distractors are chosen to be numerically close or conceptually plausible, and ambiguous items identified through repeated cross-model failures are replaced with alternatives from the same standard. The resulting benchmark is therefore designed for controlled evaluation rather than for open-ended tutoring dialogue.

### 6.3 Selected Grades, Skill Families, and Evaluation Footprint

**Grade 4** is modeled with four skill families and a four-dimensional binary mastery vector, producing $2^4 = 16$ possible skill-retention profiles.

- **S4.1**: Measurement & Data
- **S4.2**: Number & Operations (Base)
- **S4.3**: Number & Operations (Fractions)
- **S4.4**: Operations & Algebraic

**Grade 5** is modeled with three skill families in the main comparative analysis, producing a three-dimensional binary mastery vector and therefore $2^3 = 8$ possible skill-retention profiles. The broader grade-level benchmark taxonomy contains additional domains, but the main cross-model figures concentrate on the Grade-5 skill families used in the comparative analysis.

- **S5.1**: Number & Operations (Base)
- **S5.2**: Number & Operations (Fractions)
- **S5.3**: Operations & Algebraic

Across the two grades, the detailed analysis therefore covers 24 distinct target mastery profiles. Each profile corresponds to a different binary knowledge vector specifying which skills are retained and which are intentionally suppressed. The case-study visualization in Figure 5 shows six representative Grade-4 configurations for readability: the all-retained profile, the all-forgotten profile, and four mixed profiles that illustrate selective suppression patterns. The aggregate model comparisons and skill summaries, however, are computed over the full set of enumerated profiles.

In the main results narrative, the hybrid condition receives the most attention because it is the only strategy that reliably approaches the requested student profile across both analyzed grades.

Under exhaustive profile enumeration, the main two-grade hybrid comparison contains 16 Grade-4 profiles and 8 Grade-5 profiles. Evaluated across three models and 100 questions per grade, this corresponds to 7,200 model-question response events in the core comparison: $(16 \times 100 + 8 \times 100) \times 3 = 7{,}200$. This count refers only to the core hybrid analysis; the full study is larger once perfect-student baselines and prompt-strategy ablations are included.

The figures in the Results section summarize this protocol at different levels. Figure 2 isolates prompt-design effects; Figures 3-4 establish the calibration story; Figure 5 shows selective control for representative profiles; Figure 6 compares suppression difficulty across skill families; and Figure 7 closes with the aggregate retained-versus-forgotten separation.

## 7. Results

### 7.1 Prompt Strategy Comparison

All evaluated models first demonstrated strong performance on perfect-student tasks, so the central question in this section is not whether the models can solve the mathematics tasks, but whether they can be redirected toward a specified imperfect profile. We therefore begin with the prompt-ablation result that determines the rest of the paper: which prompting strategy yields the closest match between the requested skill vector and the observed behavior.

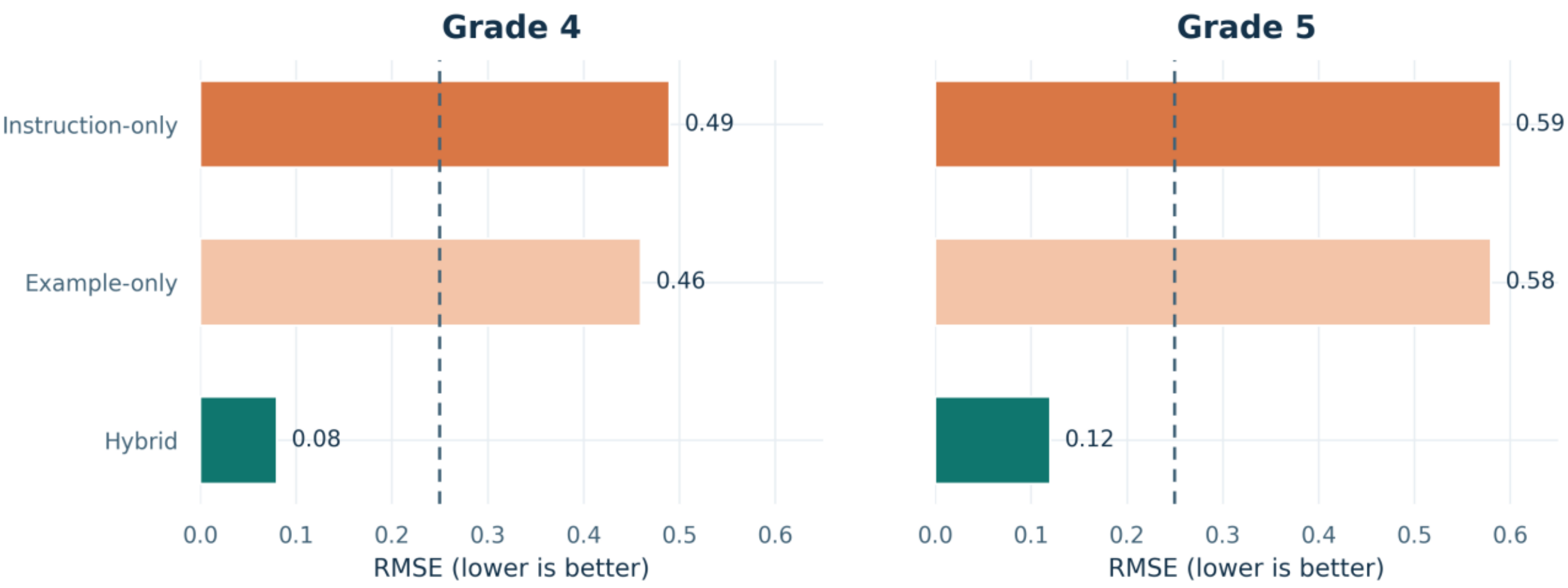

***Figure 2:*** *Prompting-strategy ablation for Grades 4 and 5. Hybrid prompting dramatically reduces RMSE compared with either instruction-only or example-only prompting, indicating that explicit constraints and demonstrations are both needed to achieve the target mastery profile. The dashed line marks a chance-level error reference under a binary target profile.*

The pattern is consistent across both analyzed grades. The hybrid strategy reaches RMSE values of 0.08 in Grade 4 and 0.12 in Grade 5, whereas example-only and instruction-only prompting remain in the 0.46-0.59 range. Because RMSE measures the average mismatch between the requested binary mastery vector and the observed skill-accuracy profile, this gap means that hybrid prompting produces answers much closer to the intended retained-forgotten pattern. Prompt design is therefore not a cosmetic detail in this setting, but a first-order determinant of controllability.

## 7.2 Cross-Skill Calibration

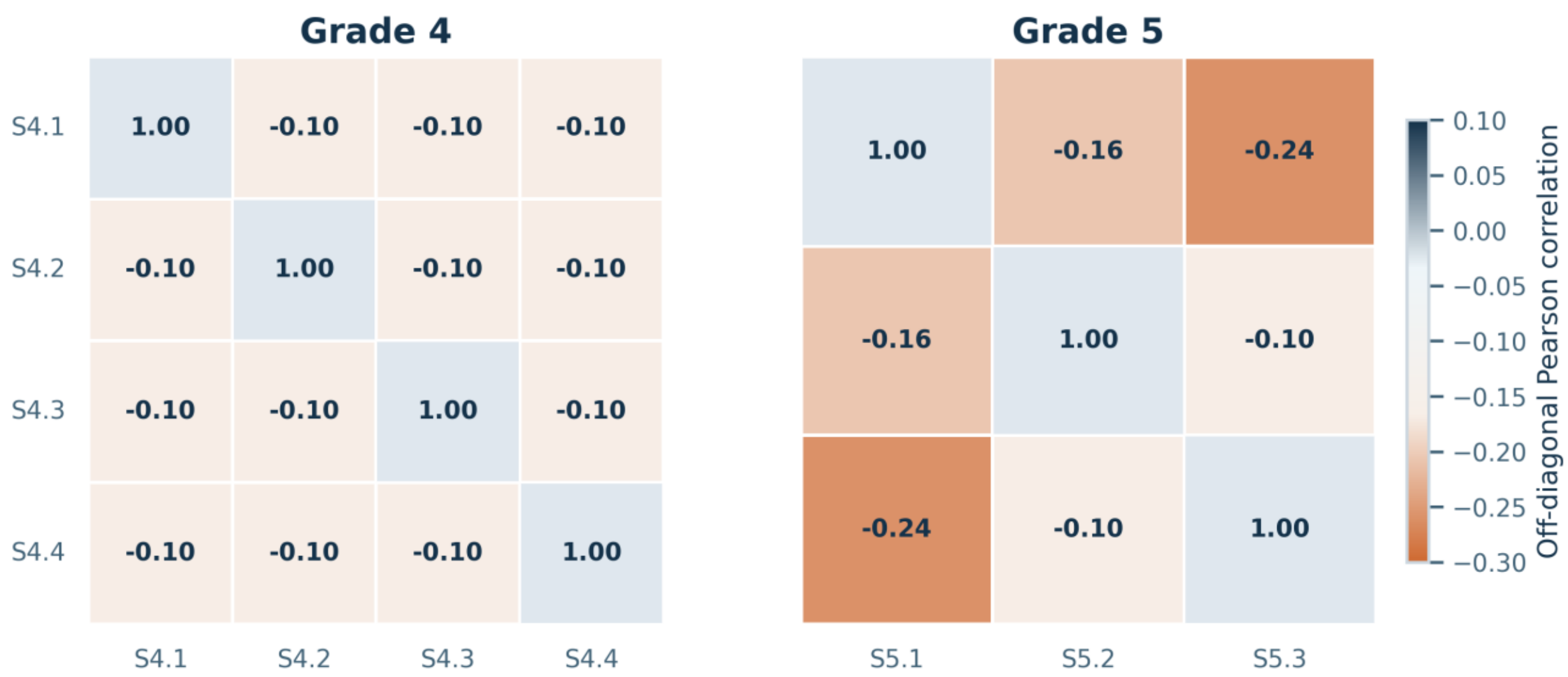

***Figure 3:*** *Skill-correlation matrices for the analyzed grades. Off-diagonal correlations remain weak, with Grade 4 clustered around -0.10 and Grade 5 ranging from -0.10 to -0.24. This supports the calibration assumption that targeted suppression can remain mostly localized rather than inducing broad failure across the skill set.*

We next ask whether the benchmark structure itself permits localized control. Figure 4 shows that cross-skill dependencies remain weak across both analyzed grades. The Grade-4 matrix is nearly uniform away from the diagonal, and even Grade 5 shows only mild negative coupling. In practical terms, the benchmark does not behave as if forgetting one skill automatically drags all other skills down with it. That said, the Grade-5 correlations are somewhat stronger than those in Grade 4, so the paper should frame independence as an empirical tendency in the setting analyzed, rather than a guaranteed property of arbitrary skill taxonomies.

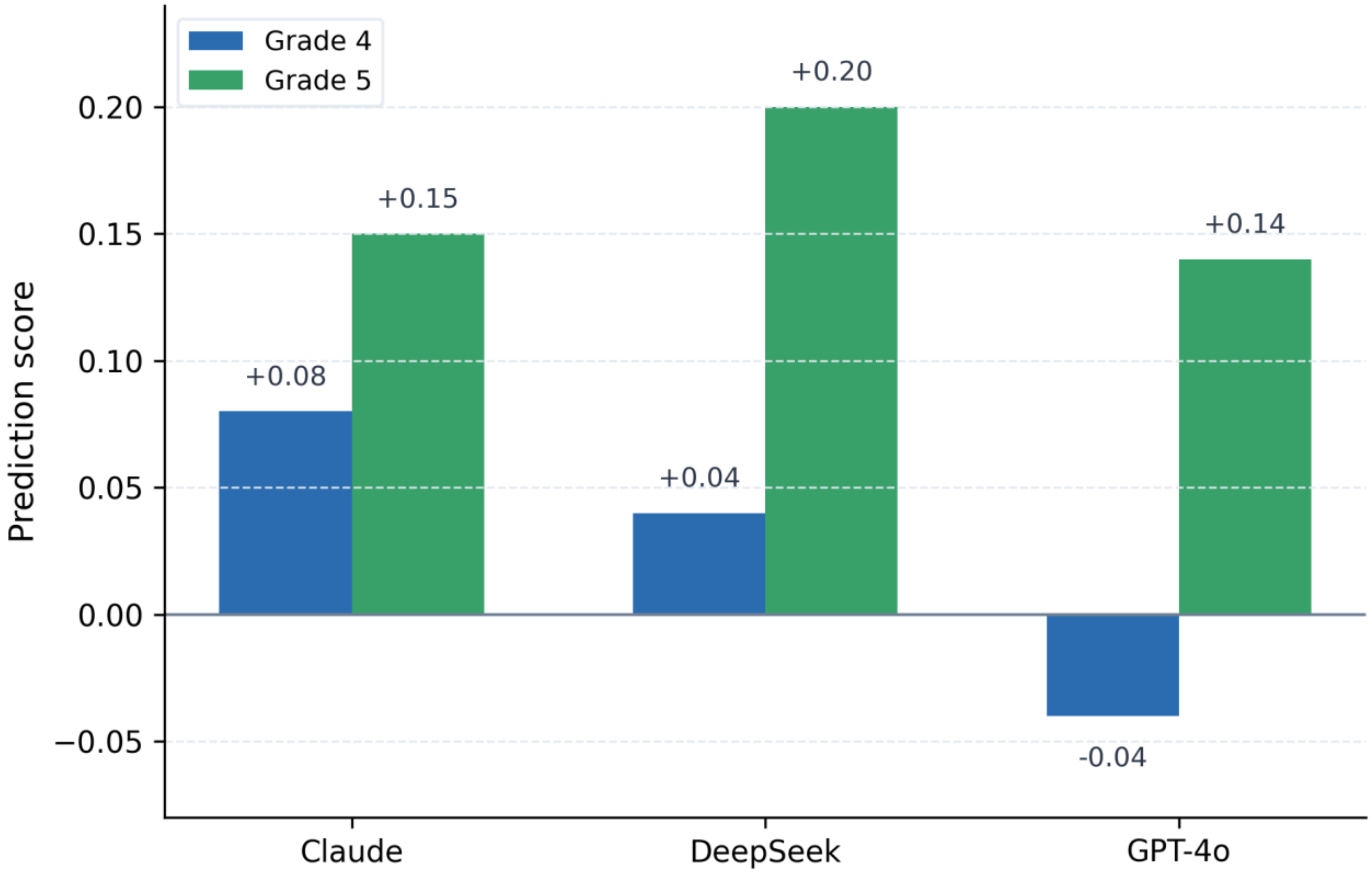

***Figure 4:*** *Prediction score by model and grade, defined as actual minus expected retained-skill accuracy. Values near zero indicate close agreement with the calibration target; positive values indicate stronger-than-expected preservation of retained skill; and negative values indicate unintended degradation.*

The calibration view summarizes the expected-versus-observed comparison of retained skill at the model level. Using the multivariate calibration procedure from Section 5.3, Figure 4 places the Grade-4 mean prediction score at +0.08 for Claude, +0.04 for DeepSeek, and -0.04 for GPT-4o. The Grade-5 panel is more favorable overall, with all three models moving into positive territory at +0.15 for Claude, +0.20 for DeepSeek, and +0.14 for GPT-4o. In concrete terms, positive values indicate that retained skills were preserved slightly better than the calibration model predicted after targeted forgetting. In contrast, the negative GPT-4o value in Grade 4 indicates a degradation of retained skills beyond expectations. Together, Figures 3 and 4 show that weak cross-skill coupling does not automatically guarantee faithful preservation of retained skill; models still differ substantially in how closely they realize the calibration target.

### 7.3 Profile-Level Selective Control

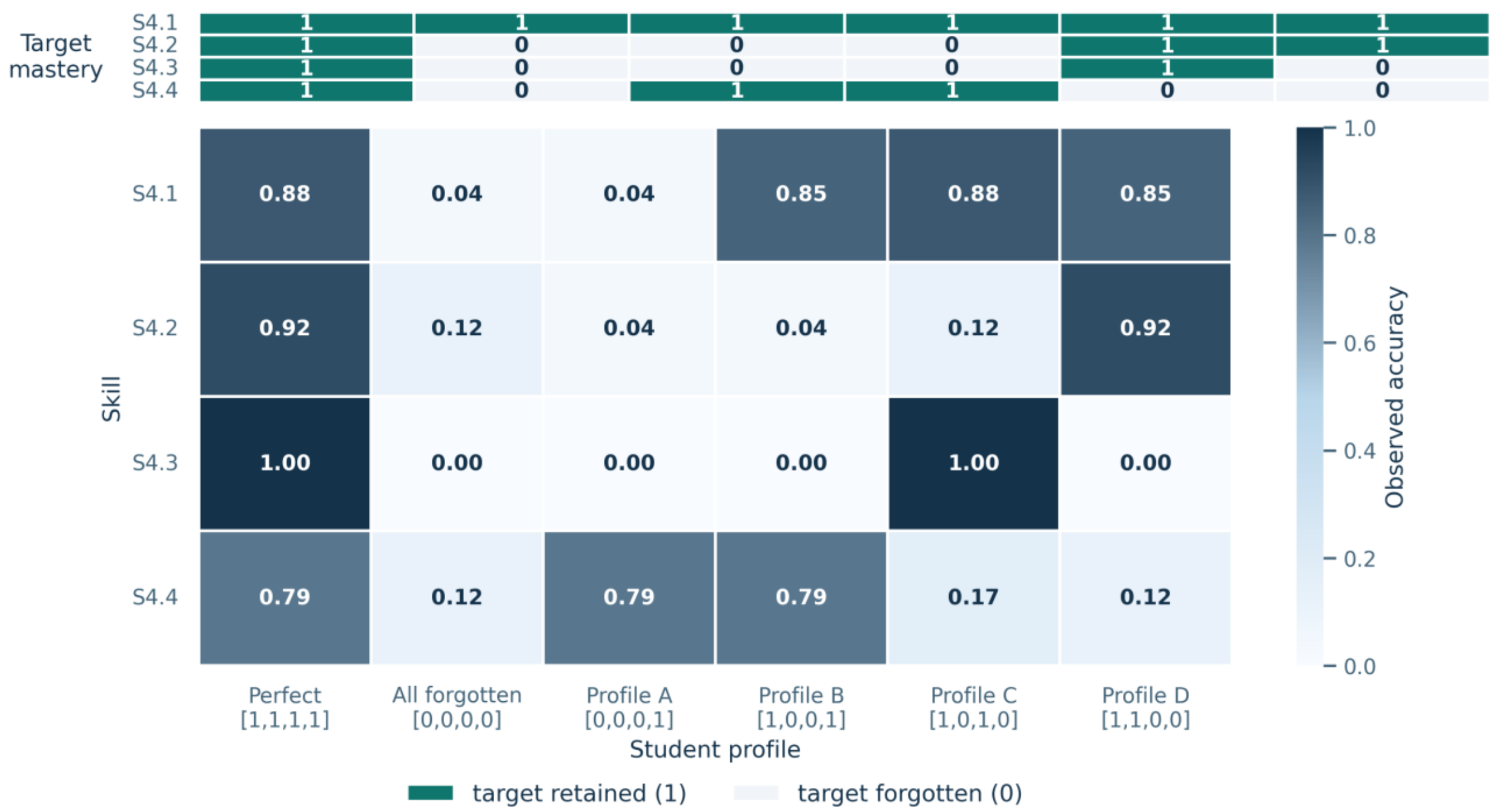

***Figure 5:*** *Grade-4 case study for Claude showing requested target mastery (top strip) and observed skill accuracy (bottom heatmap) across six representative knowledge-vector configurations. Cells near 1.00 align with retained targets, whereas cells near 0.00 align with forgotten targets. The figure highlights that profile-level control is selective rather than globally degrading.*

With the prompt strategy and calibration conditions established, Figure 5 makes the controllability phenomenon concrete at the profile level. Moving from the perfect profile to the all-forgotten profile collapses all four measured skills, while mixed profiles suppress only the intended dimensions. For example, the profile [1,0,0,1] preserves S4.1 (0.85) and S4.4 (0.79) while sharply reducing the two targeted intermediate skills. This is the clearest single-figure demonstration that the system is not merely becoming weaker overall; it is following the requested profile structure at the skill level.

### 7.4 Which Skills Resist Suppression?

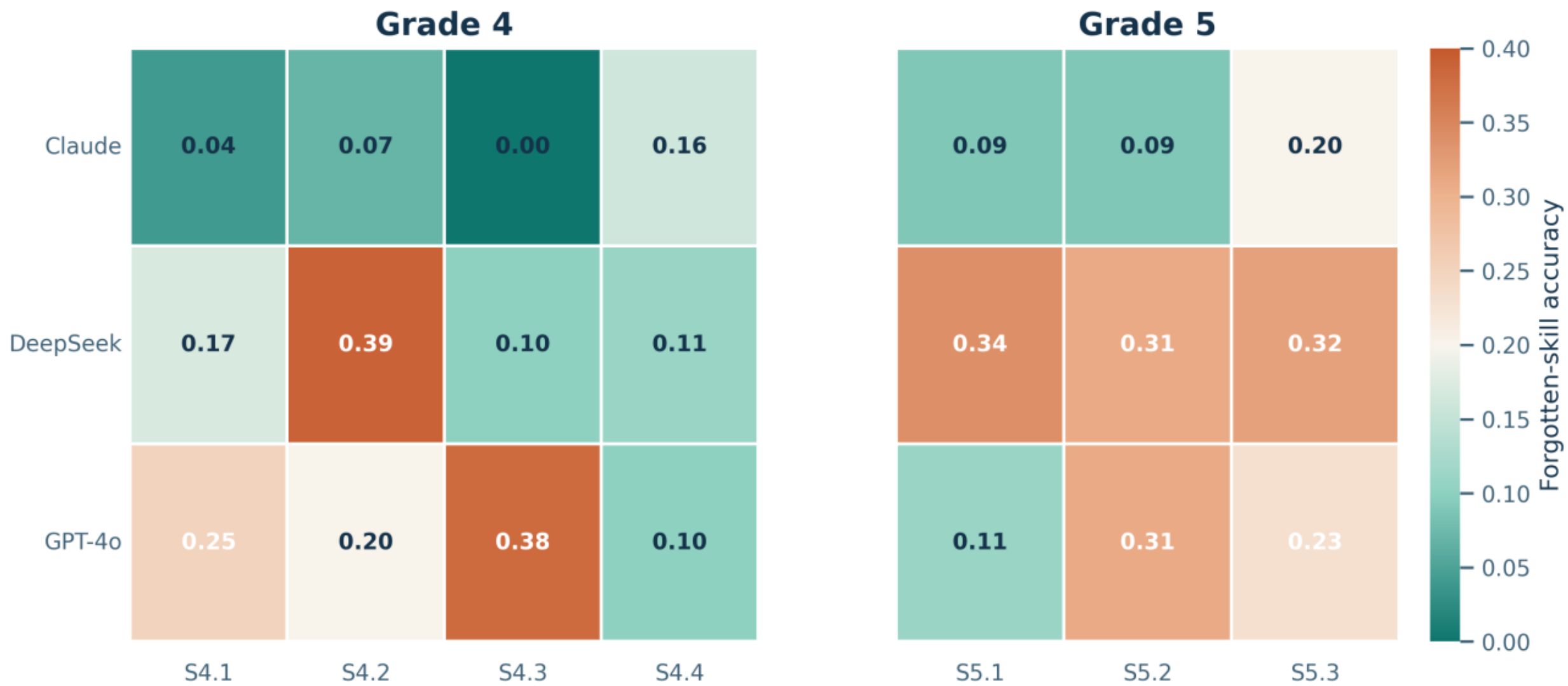

***Figure 6:*** *Forgotten-skill accuracy by model and domain. Lower values indicate stronger suppression when a skill is designated as missing in the target profile. Claude consistently occupies the lowest range, whereas DeepSeek retains substantial residual performance, and GPT-4o shows a mixed pattern that depends on the skill family.*

The domain-level view reveals where controllability succeeds and where it weakens. Claude is uniformly low across both grades, including a complete collapse to 0.00 on S4.3. DeepSeek remains substantially higher in every Grade-5 skill and on S4.2. GPT-4o is more uneven: it suppresses some domains strongly but remains relatively high on S4.3 (0.38) and S5.2 (0.31). These differences suggest that the controllability problem is shaped by both architecture and skill family rather than by a single global "obedience" factor.

## 7.5 Retained Versus Forgotten Skills

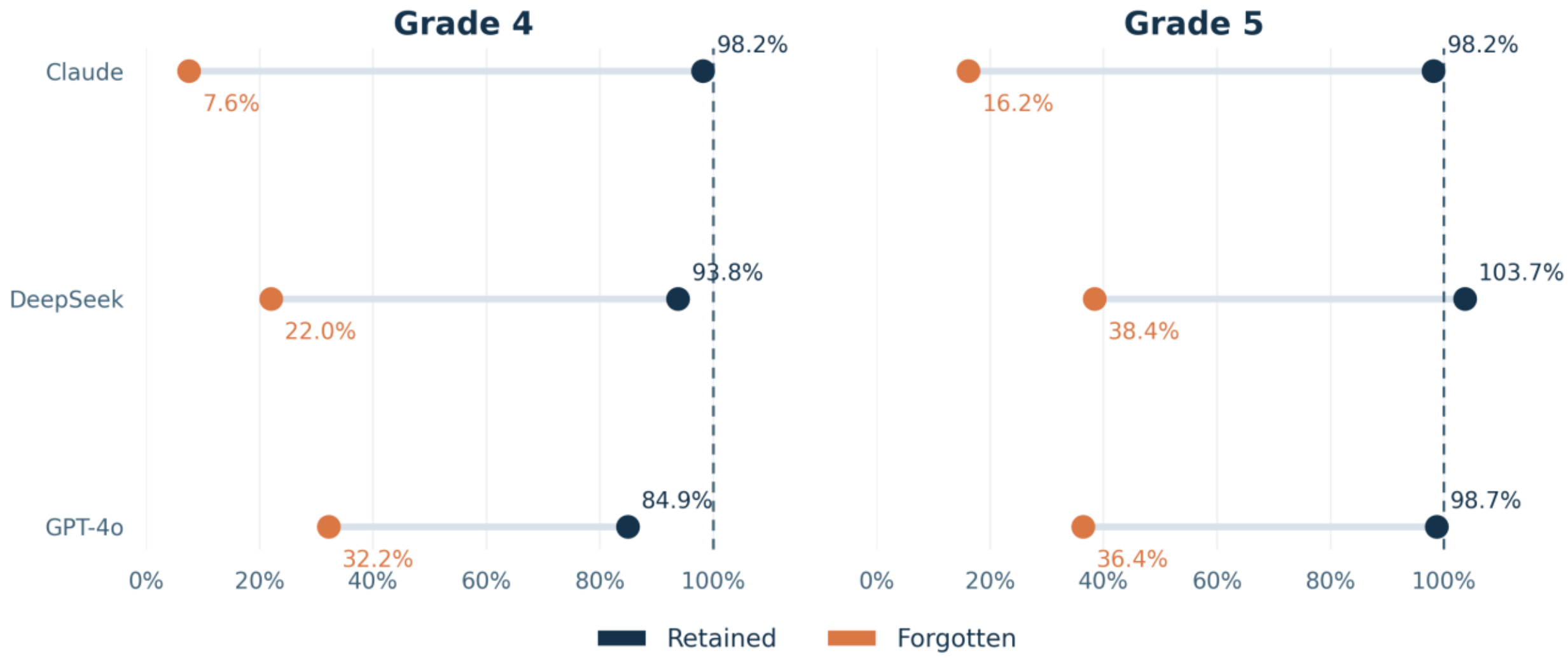

***Figure 7:*** *Relative retained-skill and forgotten-skill performance by model for Grades 4 and 5. Values are normalized to each model's perfect-student baseline, and the dashed vertical reference marks 100% of that baseline. Larger gaps indicate stronger selective forgetting with limited collateral damage to retained skills.*

The final aggregate comparison brings the main practical result into focus. Figure 7 shows a clear separation between retained and forgotten skills for every model. Claude shows the largest gap in both grades, reaching 7.6% forgotten-skill performance in Grade 4 and 16.2% in Grade 5, while keeping retained skills near baseline. DeepSeek and GPT-4o also separate retained from forgotten skills, but much less sharply. The Grade-5 retained value above 100% for DeepSeek reflects normalization against the perfect-student baseline for that same model and grade: it means the retained-skill condition slightly exceeded the measured baseline on that aggregate ratio, not that the system surpassed the underlying task definition.

## 7.6 Summary of Findings

Taken together, Figures 2-7 support six main empirical conclusions:

1. **Hybrid prompting is necessary for controllability:** the RMSE gap in Figure 2 shows that neither instruction-only nor example-only prompting is sufficient for close alignment to the target profile.
2. **Cross-skill interference remains limited in the analyzed grades:** Figure 3 shows weak off-diagonal correlations, supporting the claim that forgetting is mostly localized within this benchmark.
3. **Retained-skill calibration remains model-dependent:** Figure 4 shows that Claude and DeepSeek track the retained-skill expectation more closely, while GPT-4o exhibits larger calibration error, especially under the harder Grade-4 setting.
4. **Selective control is visible at the profile level:** Figure 5 shows that targeted skills can collapse while untargeted skills remain comparatively intact within the same simulated student profile.
5. **Suppression difficulty is skill-dependent:** Figure 6 shows that some domains, especially number-operation and fraction-related skills, are harder to suppress for certain models.
6. **Retention is substantially more stable than forgetting:** Figure 7 shows large retained-forgotten gaps for all models, with the strongest separation achieved by Claude.

## 8. Discussion

### 8.1 Educational Interpretation and Scope

The main educational implication of this work is cautious rather than expansive. The benchmark shows that an LLM can be pushed toward selective partial mastery under explicit skill constraints, a useful prerequisite for simulating student behavior. That matters for teacher-training applications because instructional practice depends on exposure to learners who are not uniformly correct.

At the same time, the present study does not validate a full classroom-facing student simulator. The evaluation operates on multiple-choice answer selection rather than on tutoring dialogue, explanation quality, misconception narratives, or teacher-student interaction. The strongest defensible interpretation is therefore that prompt-based control can produce measurable answer-level partial mastery in a benchmark setting, rather than resulting in agents that already behave like pedagogically complete students.

### 8.2 Controllability is Model-Dependent

The reported benchmark results indicate that controllability varies substantially across model families. Claude demonstrates the strongest controllability on the reported measures, DeepSeek shows greater resistance despite strong baseline performance, and GPT-4o occupies an intermediate position. High baseline mathematical competence, therefore, does not, by itself, guarantee strong controllability in this benchmark.

### 8.3 Prompt Design and Calibration Matter

Hybrid prompting substantially outperforms the individual approaches. Instructions define the desired behavior explicitly, while examples demonstrate how that behavior should appear in practice. Their combination, therefore, creates a complementary effect that is not achieved by either component alone.

The calibration results add a second lesson: weak cross-skill coupling is helpful but not sufficient. Even when the benchmark structure supports fairly localized control, models still differ in whether retained skills remain near their expected level. This is why the paper treats prompting quality, retained-skill calibration, and retained-versus-forgotten separation as complementary diagnostics rather than interchangeable ones.

### 8.4 Localized Control is Plausible but Bounded

As demonstrated by the correlation analysis, inter-skill correlations are weak in the grades analyzed, consistent with fairly localized behavioral control. The multivariate calibration framework provides an analytical interpretation of this pattern, but it should be understood as an approximation rather than as a validated cognitive model of student skill performance. Localized control appears feasible in the analyzed grades, yet it should not be assumed to transfer unchanged to larger skill graphs, richer pedagogical tasks, or alternative curricula.

Despite these encouraging findings, several constraints shape the scope and generalizability of our contributions.

### 8.5 Limitations

- **Grades analyzed in depth:** Although the benchmark was constructed for grades 1-8, the detailed controllability analyses reported here focus on Grades 4 and 5.

- **Construct validity:** Alignment to a target skill vector should not be equated with full pedagogical realism, misconception fidelity, or authentic classroom interaction.
- **Binary mastery assumption:** Oversimplifies continuous knowledge states. Future work should explore continuous mastery scales (0-1) to better reflect gradual learning progress.
- **Single-skill per question:** May not reflect real educational content where problems often require multiple skills. Hierarchical skill representations could address this.
- **Answer-only evaluation:** The current benchmark evaluates selected answers rather than full explanations or tutoring dialogue, limiting conclusions about pedagogical realism.
- **Prompt-based control only:** Fine-tuning approaches remain unexplored and may yield stronger behavioral control.
- **Mathematical domain:** Generalizability to other subjects (science, language arts, programming) untested.
- **Computational cost:** Full enumeration of $2^K$ skill configurations scales exponentially with skill count. For large-scale deployments, sampling-based approaches may be necessary.
- **Item revision procedure:** Replacing items after repeated cross-model errors may improve surface quality but can also introduce benchmark-selection bias if not governed by a separately archived audit trail.
- **Skill difficulty:** Certain skills may exhibit higher resistance to forgetting prompts than others. Systematic investigation of skill-level difficulty factors remains future work.
- **Educational misconceptions:** Current framework simulates targeted skill gaps but does not model misconception-specific reasoning patterns. Extending the framework toward misconception-grounded profiles would improve pedagogical realism.
- **Model-family reporting:** The preserved evidence supports comparison at the model-family level, but it does not fully document transient dated API snapshots for all systems.

## 9. Conclusion

This paper introduced a benchmark-oriented framework for controllable simulation of imperfect students with large language models. By representing student knowledge with explicit binary skill vectors and evaluating behavior at the skill level, the study shows that selective retention and suppression can be measured in a principled way within a mathematics benchmark setting. Across the reported experiments, combined prompting provides the most reliable alignment between the target mastery profile and observed model behavior, although the strength of controllability remains strongly model-dependent.
**Broader Impact.** This work supports a benchmark-level study of controllable imperfect-student behavior and may inform future teacher-training simulations. Potential misuse includes deploying systems that appear more pedagogically valid than they are, so any classroom-facing use should remain subject to human oversight, domain review, and explicit validity testing.

### 9.1 Future Directions

Several directions follow naturally from this work. Hierarchical skill representations could better capture the nested structure of mathematical competencies, and continuous mastery scales could reflect the gradual character of learning more faithfully than binary profiles. Beyond prompting alone, training-time or reinforcement-based methods may yield stronger and more stable controllability. Extending the framework to additional subject areas, including science, language arts, and programming, will be necessary for broader educational use. Finally, classroom-facing validation with practicing teachers remains essential for establishing practical value in authentic teacher education settings.